\documentclass[journal]{IEEEtran}
\ifCLASSINFOpdf
  \usepackage[pdftex]{graphicx}
  \graphicspath{{../pdf/}{../jpeg/}}
  \DeclareGraphicsExtensions{.pdf,.jpeg,.png}
\else
  \usepackage[dvips]{graphicx}
  \graphicspath{{../eps/}}
  \DeclareGraphicsExtensions{.eps}
\fi

\usepackage{cite}
\usepackage{amsmath,amssymb,amsfonts}
\usepackage{algorithm}
\usepackage{algorithmic}
\usepackage{url}

\begin{document}

\title{A new approach to descriptors generation for image retrieval by analyzing activations of deep neural network layers}

\author{Paweł~Staszewski, Maciej~Jaworski, Jinde~Cao,~\IEEEmembership{Fellow,~IEEE} and~Leszek~Rutkowski,~\IEEEmembership{Fellow,~IEEE}
        \thanks {P. Staszewski and M. Jaworski are with the Institute of Computational Intelligence, Czestochowa University of Technology, ul. Armii Krajowej 36, 42-200 Czestochowa, Poland, (e-mail: pawel.staszewski@pcz.pl, maciej.jaworski@pcz.pl)\protect}
        \thanks {J. Cao is with the Research Center for Complex Systems and Network Sciences, Southeast University, 210096 Nanjing, China (e-mail: jdcao@seu.edu.cn)\protect}
        \thanks{L. Rutkowski is with the Institute of Computational Intelligence, Czestochowa University of Technology, ul. Armii Krajowej 36, 42-200 Czestochowa, Poland, and also with the Information Technology Institute, University of Social Sciences, 90-113 {\L}\'o{d}\'z, Poland (e-mail: leszek.rutkowski@pcz.pl)\protect}
\thanks{This work was supported by the Polish National Science Centre under grant no. 2017/27/B/ST6/02852}}

%
%

\markboth{Journal of \LaTeX\ Class Files,~Vol.~14, No.~8, August~2015}%
{Shell \MakeLowercase{\textit{et al.}}: Bare Demo of IEEEtran.cls for IEEE Journals}
%



\maketitle

\begin{abstract}
In this paper, we consider the problem of descriptors construction for the task of content-based image retrieval using deep neural networks. The idea of neural codes, based on fully connected layers activations, is extended by incorporating the information contained in convolutional layers. It is known that the total number of neurons in the convolutional part of the network is large and the majority of them have little influence on the final classification decision. Therefore, in the paper we propose a novel algorithm that allows us to extract the most significant neuron activations and utilize this information to construct effective descriptors. The descriptors consisting of values taken from both the fully connected and convolutional layers perfectly represent the whole image content. The images retrieved using these descriptors match semantically very well to the query image, and also they are similar in other secondary image characteristics, like background, textures or color distribution. These features of the proposed descriptors are verified experimentally based on the IMAGENET1M dataset using the VGG16 neural network.
\end{abstract}


%
\IEEEpeerreviewmaketitle

\section{Introduction}
Content-based image retrieval (CBIR) is a very important and challenging issue applicable in many scientific and business fields of human activity. Among others, these are: automotive industry, medical industry, virtual reality, games, retail, security industry, social media platforms, visual search engines, and many more. Each of these application areas requires a different approach to the analysis of image content. Depending on the aim of processing, we can analyze in detail the objects on the image, as well as surroundings, background textures, foreground elements of the stage, etc. Therefore, CBIR is a very broad field of computer science research. Nevertheless, regardless of the application aim and the chosen method, efficiency always plays the most important role.

In the literature there are many methods that allow achieving satisfactory effectiveness for CBIR task \cite{ismail2017survey}\cite{gu2019survey}\cite{zhou2017recent}\cite{chauhan2019efficient}\cite{dubey2016interactive}. Most of them are based on the general scheme of descriptors construction, using techniques for extracting the most characteristic features contained in the image \cite {radenovic2016cnn} \cite {somasundaran2020robust} \cite {wu2019multi} \cite {saritha2019content} \cite {long2019deep} \cite{zhou2018region}. It is also worth paying attention to methods that allow constructing effective descriptors for human faces comparison. They are successfully used in many business areas \cite {jang2018deep} \cite {pham2019facial}.

One of the interesting and effective approaches to descriptor generation is the analysis of neural activations in deep models for image classification. Currently, the state-of-art neural networks for image processing are convolutional neural networks. The majority of them, like the VGG16 net applied in this paper, contain additionally one or two fully connected layers placed on top of the last convolutional block for classification purposes. In \cite{babenko2014neural} the descriptors called neural codes were proposed, which are composed of the signals of neurons in the fully connected layers. Such descriptors turned out to be very effective. However, activations of fully connected layers contain the information required to perform classification on images. This is desirable when the aim is to retrieve images which contain objects of the same class as a query image.  It is obviously a commonly encountered task in various fields of research or applications, however, in some cases, it might be not enough. Some tasks may require to store in descriptors not only information about the object classes but also about the background, surrounding or textures. To this end, the information from convolutional layers should be somehow incorporated into descriptors. Convolution neural networks play a very important role in CBIR. They allow the extraction of very detailed features that can be used to construct an effective descriptor \cite {jose2018pyramid} \cite {dong2019multilayer} \cite {tzelepi2018deep} \cite {jun2019combination}. Although the information contained in activations of convolutional layers is highly useful in tasks like denoising, segmentation, and classification of images,  they are at the same time hardly interpretable. A large number of connections in deep neural structures and complex dependencies between neuron activations makes it difficult to separate only those features that are most important for a given image. Unfortunately, existing solutions do not provide the perfect extraction of features. Motivated by this fact, in this paper we will develop a new type of descriptors, extending the idea of neural codes presented in \cite{babenko2014neural} by concatenating the most significant activations from convolutional layers with those from fully connected ones. Such descriptors separate irrelevant noise from valuable knowledge describing the image. It should be noted that a similar approach was previously applied to the generation of hashes, i.e. the binary descriptors \cite{Wu2017}, \cite{Ng2020}, \cite{Jin2020}. The binary codes can be very useful in retrieving images from very large databases, however, they cannot guarantee as high accuracy as real-valued descriptors. Therefore, in this paper we focus on descriptors which values are taken directly from neural activations. In the following sections we will present our novel  algorithm that analyzes the blocks of convolutional layers of the neural network and selects the signals most relevant from the classifier's decision perspective. Because convolution layers are built on the basis of filters, their activities will be stronger for features enhancing the texture of the image. By concatenating such activations from the convolutional part of the network with the values collected from the signals of fully connected layers, we obtain a set of features that focus on the semantic category of objects contained in the image as well as other mentioned previously characteristics of the image, like textures or backgrounds. The generated descriptor will contain very detailed features describing the entire image content. In this paper, we use the commonly known VGG16 neural network because of its relatively simple structure. However, our method for generating descriptors can be adapted to networks with other structures straightforwardly. 

The significance, characteristics, and originality of this paper, relative to the state-of-art, are given as follows:
\begin{itemize}
\item[1)] A novel method for content-based image retrieval, based on neural codes, is proposed.
\item[2)] The image descriptors' construction is based on the synergy of the spatial information from the convolutional layers with the information represented by the two fully connected layers -- commonly used for classification tasks.
\item[3)] The presented approach, as it is depicted in the series of experiments, perfectly represents the content image by preserving the semantic similarity and, simultaneously, the structural similarity,
\item[4)] The method, contrary to the previous attempts in the literature, is not demanding from the computational point of view, and, moreover, it can be extended in many directions, as we suggest in Section \ref{sec5}.
\end{itemize}

The rest of this paper is organized as follows. Section \ref{sec2} presents a description of the VGG16 neural network and introduces a notation useful in describing our algorithm. In Section \ref{sec3} a step by step explanation of our method is provided. Experimental results demonstrating the effectiveness of the proposed approach are presented in Section \ref{sec4}. Section \ref{sec5} concludes the paper and outlines some possible directions for future research.

\section{Preliminaries -- The VGG16 net structure}
\label{sec2}
\begin{figure}[!t]
\centering
\includegraphics[width=3.0in]{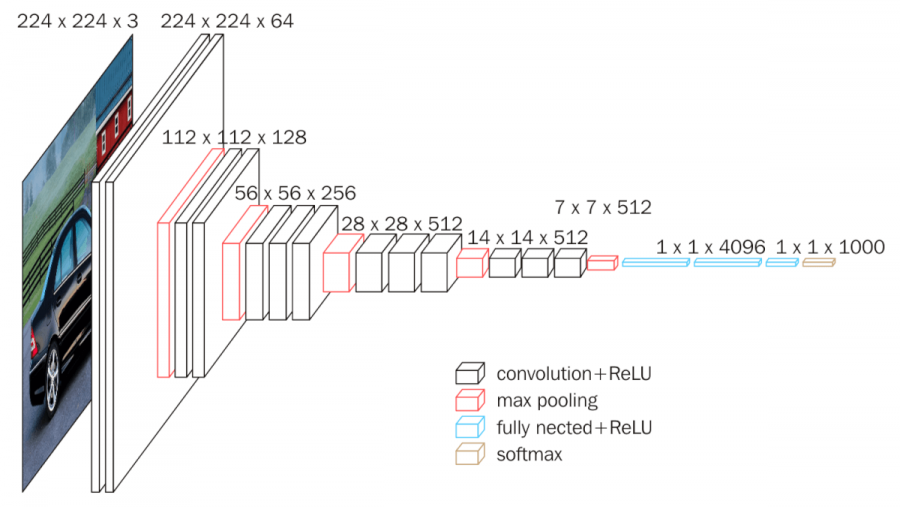}
\caption{The VGG16 structure.}
\label{fig:vgg16_structure}
\end{figure}

It should be pointed out that our method for descriptors construction can be applied for deep neural networks of various structures. However, in this paper, we focused our analysis on the VGG16 net \cite{qassim2018compressed} because of its simplicity. We chose this network since its layer-wise architecture allows us to understand the relations between neuron activations relatively easily. The structure of the VGG net is visualized in Fig. \ref{fig:vgg16_structure}. 

Let us first introduce some useful notations to describe the VGG network. To make the description as clear as possible, we focus only on these elements, which are really required to present how our approach works. As can be seen in Fig. \ref{fig:vgg16_structure} the VGG16 net is composed of a block of layers. Each block contains two or three convolutional layers, which are followed by a max-pooling layer. For clarity, in the text, we will refer to this block of convolutional and max-pooling layers simply as blocks. Let $L_{1},\dots,L_{N}$ denote the subsequent blocks of the network and let $y_{1},\dots,y_{N}$ be their outputs, i.e.
\begin{equation}
    \label{eq2}
    y_{m} = L_{m}(y_{m-1}) = L_{m} \circ \dots \circ L_{1}(x),\; m=1,\dots,N.
\end{equation}
where $x \equiv y_{0}$ is the input vector. Strictly speaking, $y_{m}$ is the output of the max-pooling layer placed at the end of the $m$-th block of convolutional layers. The output of the $m$-th block consists of $C_{m}$ feature maps (also called channels) being of the size $W_{m}\times H_{m}$. Knowing that the ReLU function is used as an activation function in each layer, the domain of possible values of the $m$-th block is given by

\begin{equation}
    \label{eq3}
    y_{m} \in \mathbb{R}_{+}^{W_{m}\times H_{m}\times C_{m}}
\end{equation}
Hence, each $y_{m}$ is a rank-$3$ tensor, and it consists of $C_{m}$ rank-2 tensors

\begin{equation}
    \label{eq4}
    y_{m} = \left(y_{m,1},\dots,y_{m,C_{m}}\right),\; y_{m,k} \in \mathbb{R}_{+}^{W_{m}\times H_{m}}
\end{equation}
for $k = 1,\dots,C_{m}$. Values of $W_{m}$, $H_{m}$ and $C_{m}$ for subsequent blocks can be easily read out from Fig. \ref{fig:vgg16_structure}. 

On top of the last (i.e. the $N$-th) block of convolutional layers, there are two fully connected layers

\begin{eqnarray}
    \label{eq5}
    y_{N+1} = FC_{1}(y_{N}), y_{N+1} \in \mathbb{R}_{+}^{D_{1}},\\\nonumber
    y_{N+2}= FC_{2}(y_{N+1}), y_{N+2} \in \mathbb{R}_{+}^{D_{2}}.
\end{eqnarray}
consisting of $D_{1}$ and $D_{2}$ neurons, respectively ($D_{1}=D_{2} = 4096$ in this case). The fully connected layers are followed by a softmax one

\begin{equation}
\label{eq6}
    y = S(y_{N+2}).
\end{equation}
Summarizing, in the presented notations the whole neural network realizes the following function 

\begin{equation}
\label{eq1}
    y = S \circ FC_{2} \circ FC_{1} \circ L_{N} \circ L_{N-1} \circ \dots \circ L_{2} \circ L_{1}(x).
\end{equation}

\section{A novel method for image descriptors generation}
\label{sec3}

A standard method for building image descriptors using the convolutional neural network is to take, for each image $x$, the activations obtained from fully connected layers. Such descriptors are known in the literature under the name of neural codes \cite{babenko2014neural}.  In the case of the VGG16 net, one could take activations from one of the fully connected layers or a concatenation of activations from both of them. In this work, we consider the latter approach. Let us denote such FC-based descriptor as $h(x)$

\begin{equation}
    \label{eq7}
    h(x) =\left[ y_{N+1},y_{N+2}\right]_{x},
\end{equation}
where $y_{N+1}$ and $y_{N+2}$ are defined by (\ref{eq5}). The subscript $x$ denotes that the neural activations used in the descriptor were obtained for data element $x$. Hence, in this case the descriptor is a $\left(D_{1}+D_{2}\right)$-dimensional vector. The fully connected layers contain the compressed information from convolutional layers, consisting of features useful for the image classification task. Therefore, the image retrieval based on such descriptors gives satisfactory results only if the aim is to search for images of the same class as the query image. Alternatively, there are many other tasks of widely understood information retrieval. We might be interested not only in searching for objects of the same class but also for other features like the scenery (background), the performed activity, or the emotions captured on someones' faces. All this kind of information is omitted in the FC-based descriptors since it is contained mainly in the activations of convolutional layers. 

Based on this observation, we present an extension of the idea of neural codes by adding information from convolutional layers. However, the question arises which activations and from which convolutional layers should be actually taken into account. The total number of such neurons is very large. Taking all of them would result in very high-dimensional descriptors. Moreover, not all activations carry the same amount of information, which is important from the perspective of the final output of the neural network. Hence, the best way seems to be to neglect low-valued activations, i.e. to choose only these neurons which influence the final decision of the classifier the most. To keep the dimensionality of the constructed descriptors relatively low, we consider taking activations only from max-pooling layers, not from the convolutional layers preceding them.



\begin{figure}[!t]
\centering
\includegraphics[width=3.0in]{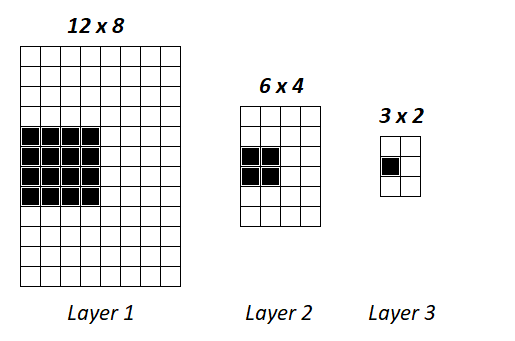}
\caption{An example of localization of neural regions with significant activations in subsequent layers based on significant activations in the last convolutional layer.}
\label{fig:upsampling_localization}
\end{figure}

Since the convolutional layers are followed by the ReLU activation function, it is reasonable to assume for further considerations that only high activations with a positive sign have an impact on the final decision of the classifier. As can be seen in Fig. \ref{fig:vgg16_structure}, the output of the last (i.e. the $N$-th) convolutional layer block is of the shape $W_{N}\times H_{N} \times C_{N} = 7 \times 7 \times 512$ -- although the number of channels is large, the feature maps sizes are small. Therefore, the analysis of signals flow is relatively easy. Each feature map in this layer consists of $W_{N}\times H_{N} = 7 \times 7 = 49$ neurons, among which only a few will presumably be activated with significantly high values. We can impose some threshold value $q$ and consider only these neurons for which the activation signal is above it (in the simulations performed in Sec. \ref{sec4} $q=0.5$ was chosen experimentally). To make the explanation of our approach more clear, let us introduce for each feature map  $y_{N,k}, k=1,\dots,C_{N}$, in the final layer of the last block a corresponding matrix $z_{N,k} \in \{0;1\}^{W_{N}\times H_{N}}$

\begin{equation}
    \label{eq8}
    \left(z_{N,k}\right)_{ij} = \begin{cases}1, \left(y_{N,k}\right)_{ij} \ge q \\0, \left(y_{N,k}\right)_{ij} < q\end{cases}.
\end{equation}This matrix contains information about which neurons are significant. Having these 'significance matrices' computed for the last block of convolutional layers, the next step is to select important neurons in previous blocks. We proposed to make it in the following way. First, it should be noticed that due to max-pooling layers, the sizes of feature maps for subsequent blocks differ by half, i.e.

\begin{eqnarray}
\label{eq9}
W_{m} = 2W_{m-1},&\;m=N,\dots,2,\\
H_{m} = 2H_{m-1}.&
\end{eqnarray}In subsequent blocks from $m=1,\dots,(N-1)$ we assume that significant neurons are those which correspond spatially to significant neurons in the $N$-th block. This is schematically demonstrated in Fig. \ref{fig:upsampling_localization}. 

Let us assume that $i_{m} \in A_{m}=\{0,\dots,H_{m}-1\}$ and $j_{m} \in B_{m}=\{0,\dots,W_{m}-1\}$. Then, if the neuron indexed by a pair $(i_{N,k},j_{N,k})$ in the $k$-th feature map of the $N$-th block is marked as significant, then the neurons indexed by the following set of index pairs in the $m$-th block are also considered significant

\begin{eqnarray}
\label{eq10}
&I_{m,k}(i_{N,k},j_{N,k}) = \\\nonumber &\{(i,j) \in A_{m}\times B_{m}: \lfloor\frac{i}{2^{N-m}}\rfloor=i_{N,k} \land  \lfloor\frac{j}{2^{N-m}}\rfloor=j_{N,k} \},
\end{eqnarray}
where $\lfloor x \rfloor$ denotes the floor of $x$, i.e. the greatest integer less than or equal to $x$. Hence, the final set of indices of significant neurons in the $m$-th block, corresponding to the $k$-th feature map of the last $N$-th block, is given by

\begin{equation}
    \label{eq10.5}
    Z_{m,k} = \bigcup\limits_{(i,j):\left(z_{N,k}\right)_{ij}=1}I_{m,k}(i,j)
\end{equation}
Based on sets $Z_{m,l}$ obtained for all feature maps $l=1,\dots,C_{N-1}$ from the last block, a significance matrix $z_{m}$ for the $m$-th layer can be constructed in the following way

\begin{equation}
\label{eq11}
\left(z_{m}\right)_{ij}= \sum_{k=1}^{C_{N}}\chi_{Z_{N,k}}(i,j),
\end{equation}
where $\chi_{A}(X)$ is an indicator function of set $A$( i.e. it returns $1$ if $X \subset A$ and $0$ otherwise). In brief, matrix $z_{m}$ counts the number of sets $z_{N,k}$ which contain pair $(i,j)$. It should be noted that for $m = N-1,\dots,1$ the significance matrices are significantly differ from those for $m=N$. First, there is no need to consider separate significance matrices for different feature maps since all they are equal, i.e. $z_{m,k}\equiv z_{m}$, $k=1,\dots,C_{m}$, hence the second index can be omitted. Moreover, their elements can take any natural number value, not only $0$ or $1$. 
    
 To keep the dimensionality of our final descriptors fixed, we want to compute only one characteristic value for each feature map in each block. We denote this quantity as $w_{m,k}$ for the $k$-th feature map of the $m$-th block of convolutional layers. Their values are calculated in the following way
 
 \begin{equation}
 \label{eq12}
 w_{m,k} = \frac{\sum_{i=1}^{H_{m}}\sum_{j=1}^{W_{m}}\left(z_{m}\right)_{ij}\left(y_{m,k}\right)_{ij}}{\sum_{i=1}^{H_{m}}\sum_{j=1}^{W_{m}}\left(z_{m}\right)_{ij}},
 \end{equation}
where $m=1,\dots,N$ and $k=1,\dots,C_{m}$. For $m=N$, the characteristic values are simply arithmetic averages of significant neuron activities. In the case of $m=N-1,\dots,1$ the averages are weighted. The weight of the neuron from the $m$-th layer is equal to the number of significant neurons from the $N$-th block, which correspond to this neuron spatially. Let us define $w_{m} = \left[w_{m,1},\dots,w_{m,C_{m}}\right]$. Then, we propose the following two new types of descriptors, analogous to the neural code given by (\ref{eq7}):

\begin{itemize}
\item A concatenation of vectors $w_{m}$ for all blocks \begin{equation}
\label{eq13}
\tilde{\eta}(x) = \left[w_{1},\dots,w_{N}\right]_{x}.
\end{equation}
\item A concatenation of the neural code given in (\ref{eq7}) with vector $\tilde{\eta}(x)$ \begin{equation}
\label{eq14}
\eta(x) = \left[h(x),\tilde{\eta}(x)\right]=\left[y_{N+1},y_{N+2},w_{1},\dots,w_{N}\right]_{x}.
\end{equation}\end{itemize}

Dimensionality of these descriptors is equal to $\sum_{m=1}^{N}C_{m}$ for $\tilde{\eta}(x)$ and $D_{1}+D_{2}+\sum_{m=1}^{N}C_{m}$ for $\eta(x)$. In Algorithm \ref{alg1} we present the pseudocode, which summarizes the description provided above. The algorithm returns descriptors $h(x)$, $\tilde{H\eta}(x)$ and $\eta(x)$ for a given image $x$. The descriptors obtained for training data and query images can be further compared using commonly used quantities, like the $L_{1}$ distance or the Euclidean distance.

\begin{algorithm}
\caption{Descriptor construction}
\label{alg1}
\begin{algorithmic}
\REQUIRE image $x$
\ENSURE descriptors $h(x)$, $\tilde{\eta}(x)$, $\eta(x)$
\STATE Generate $y_{1} = L_{1}(x)$
\FOR{$m=2$ \TO $m=N$} \STATE {$y_{m}=L_{m}(y_{m-1})$} \ENDFOR
\STATE $y_{N+1}=FC_{1}(y_{N})$
\STATE $y_{N+2}=FC_{2}(y_{N+1})$
\FOR{$k=1$ \TO $k=C_{N}$} \STATE {compute $z_{N,k}$ using (\ref{eq8})} \ENDFOR
\FOR{$m=(N-1)$ \TO $m=1$} {\FOR{$k=1$ \TO $k=C_{N}$}\STATE {Create sets $Z_{m,k}$ using (\ref{eq10.5})} \ENDFOR
\STATE Compute $z_{m}$ using (\ref{eq11})
\FOR{$k=1$ \TO $k=C_{m}$} \STATE {Compute $w_{m,k}$ using (\ref{eq12})} \ENDFOR
\STATE $w_{m} = \left[w_{m,1},\dots,w_{m,C_{m}}\right]$
} \ENDFOR
\STATE $h(x) = \left[y_{N+1},y_{N+2}\right]_{x}$
\STATE $\tilde{\eta}(x) = \left[w_{1},\dots,w_{N}\right]_{x}$
\STATE $\eta(x) = \left[y_{N+1},y_{N+2},w_{1},\dots,w_{N}\right]_{x}$
\RETURN $h(x)$, $\tilde{\eta}(x)$, $\eta(x)$
\end{algorithmic}
\end{algorithm}

\section{Experimental results}
\label{sec4}

\subsection{Dataset description}

As it was previously mentioned, we applied the VGG16 neural network to demonstrate the performance of our method. We used the network already trained on the ILSVRC dataset \cite{russakovsky2015imagenet}, \cite{imagenet_cvpr09}. To investigate the effectiveness of descriptors discussed in this paper, we performed computations on the IMAGENET1M dataset \cite{CGW17}, which was created to conduct research in the area of image retrieval. This dataset contains images from the ILSVRC. Each image belongs to one of $1000$ classes. The total number of training data is equal to $1281167$. Additionally, a special set of $25000$ query images is also available. It should be noted that the IMAGENET1M dataset also provides its own descriptors assigned to each image. Generally speaking, these descriptors are generated by the additional hidden neuron layer placed after the last fully connected layer. This hidden layer is trained to learn effective codes for images. For details, the reader is referred to \cite{CGW17}. In this paper, we will denote this IMAGENET1M descriptor for image $x$ as $IM(x)$.

\subsection{Comparison of considered descriptors}

\begin{figure}[!t]
\centering
\includegraphics[width=3.0in]{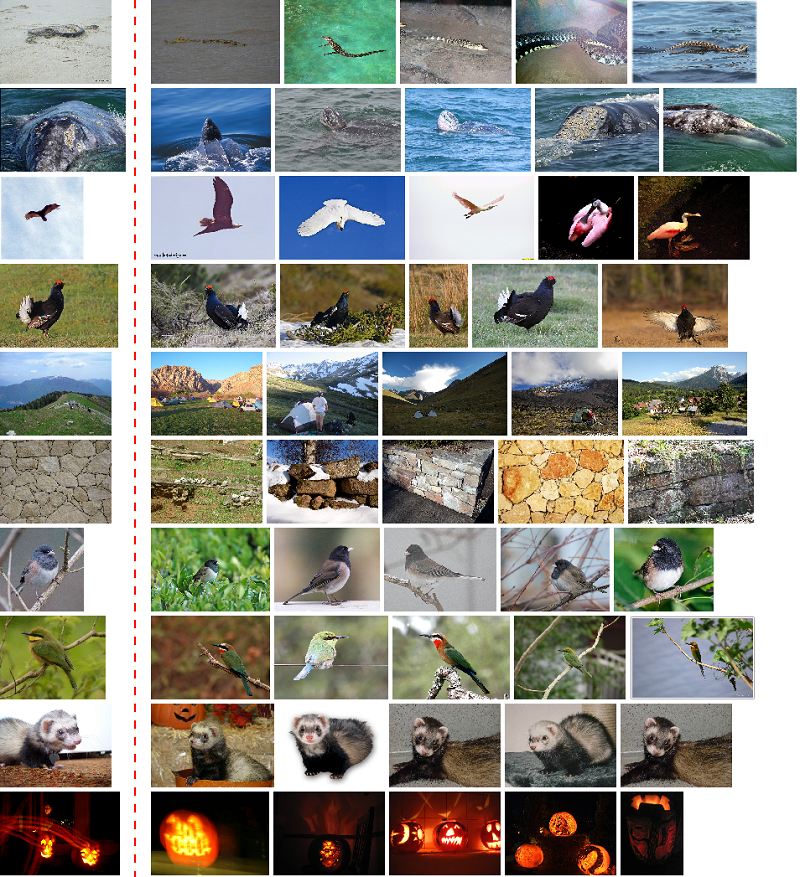}
\caption{The top-5 most similar images for ten random query images using descriptors $IM(x)$ taken directly from the IMAGENET1M database.}
\label{fig:sim_base}
\end{figure}

\begin{figure}[!t]
\centering
\includegraphics[width=3.0in]{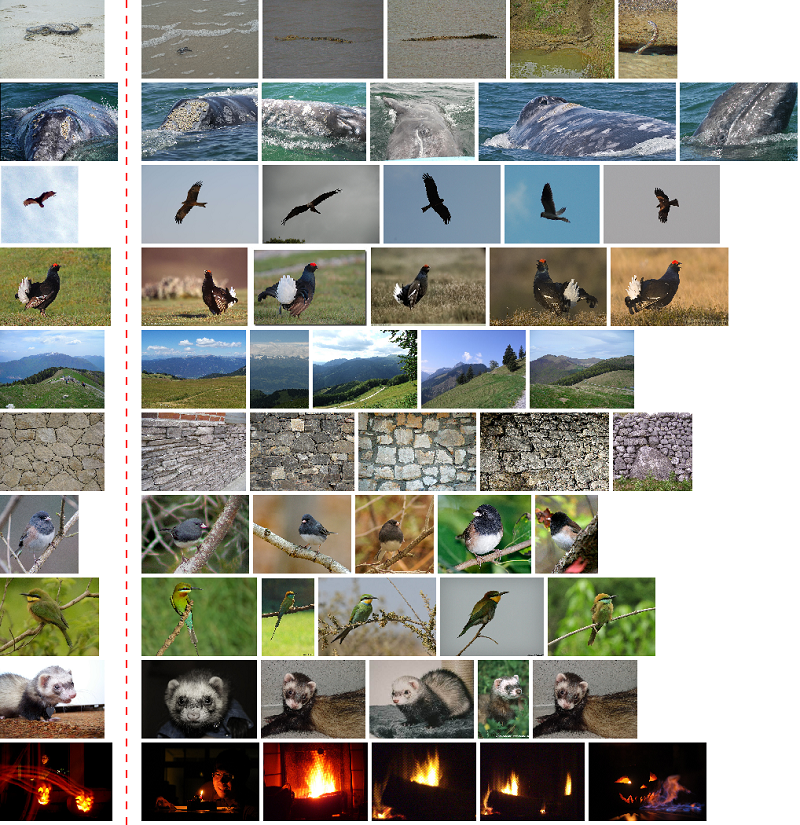}
\caption{The top5 most similar images for ten random query images using descriptors $h(x)$ based on fully connected layers.}
\label{fig:sim_fc}
\end{figure}

\begin{figure}[!t]
\centering
\includegraphics[width=3.0in]{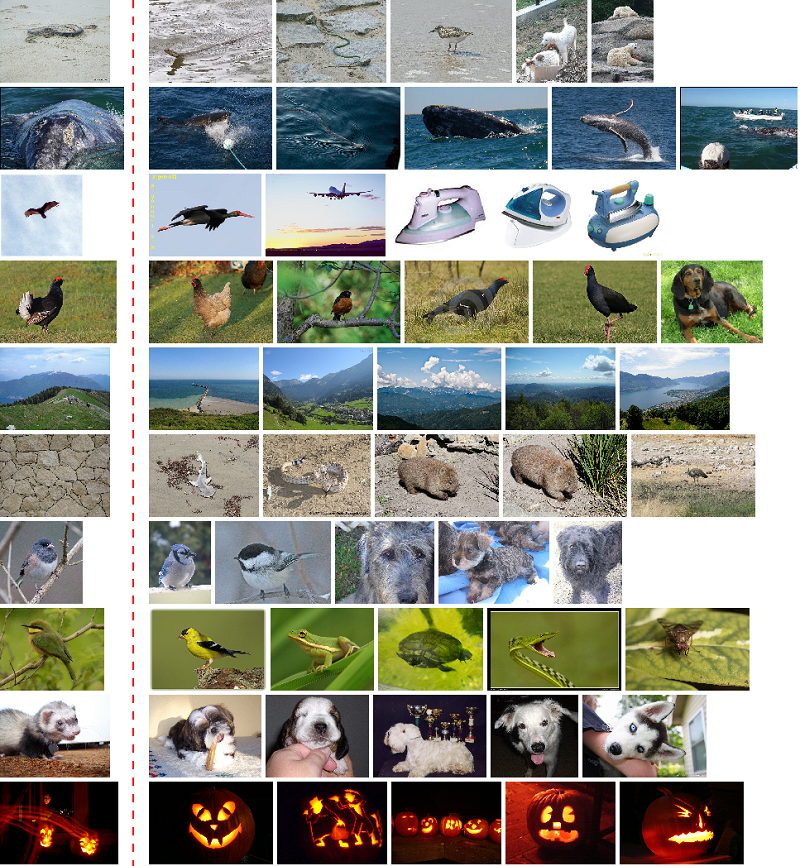}
\caption{The top-5 most similar images for ten random query images using descriptors $\tilde{\eta}(x)$ based solely on convolutional layers.}
\label{fig:sim_cnn}
\end{figure}

\begin{figure}[!t]
\centering
\includegraphics[width=3.0in]{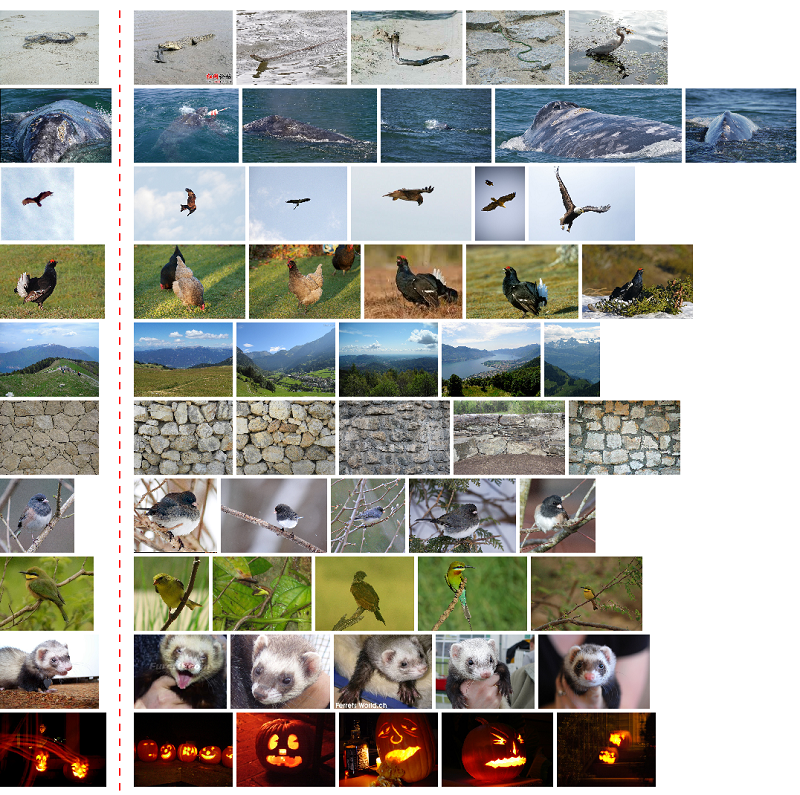}
\caption{The top-5 most similar images for ten random query images using descriptors $\eta(x)$ based on both convolutional and fully connected layers.}
\label{fig:sim_own}
\end{figure}

At the beginning, for each image $x$ from training set, we generated three descriptors $h(x)$, $\tilde{\eta}(x)$ and $\eta(x)$ using Algorithm \ref{alg1}\footnote{All the experiments were conducted using our own software implemented in Python and it can be found at \url{https://github.com/pstaszewski/cbir_2020_04}}. Additionally, for each image, there is a descriptor $IM(x)$ available in the IMAGENET1M dataset. Next, we randomly chose $10$ query images, each belonging to a different class. Then for each query image, we found top-$5$ most similar images from the training set with respect to four considered descriptors. The similarity is measured by a distance between two descriptors. We applied the $L_{1}$ distance since its computational cost is relatively low comparing to, for example, the Euclidean distance. We divide the distances by the number of dimensions to make the measure independent on the descriptor dimensionality. The results obtained for descriptors $IM(x)$, $h(x)$, $\tilde{\eta}(x)$ and $\eta(x)$ are presented in Fig. \ref{fig:sim_base}, \ref{fig:sim_fc}, \ref{fig:sim_cnn} and \ref{fig:sim_own}, respectively\footnote{For clarity of the paper we presented only top-5 most similar images for each query image. More extended results with top-10 images can be found at \url{https://github.com/pstaszewski/cbir_2020_04/tree/master/Results}}.

Obviously, the assessment of which of the images are more similar to the corresponding query image is mainly a subjective matter. Nevertheless, several objective remarks still can be made. The images found using $IM(x)$ descriptor or $h(x)$ neural code in almost every case match in the class with the query image. Hence, these descriptors surely contain information about the semantic meaning of the images. However, if we wanted the images to be similar concerning other characteristics, like textures, background or color distribution, then the obtained results would not be considered satisfactory. This is caused by the fact that information contained in descriptors constructed using the last layers of the neural network are focused only on the classes of the objects. This is contrary to the case of descriptor $\tilde{\eta}(x)$, constructed only on the basis of convolutional layers activations. Here the images found in the training set rarely agree with the class of a query image. However, the textures and colors visually are very similar. It is clearly visible for two images of birds: for the green one, the algorithm found objects like a frog or green insect, whereas for the blue bird it returned a monkey or dogs with a similar color. The advantages of both descriptors $h(x)$ and $\tilde{\eta}(x)$ are revealed in the very good performance of the descriptor $\eta(x)$, which is a combination of the two. Now, the retrieved images match semantically to the query image, and simultaneously, they also agree in other, secondary characteristics mentioned previously.

\begin{figure}[!t]
\centering
\includegraphics[width=3.0in]{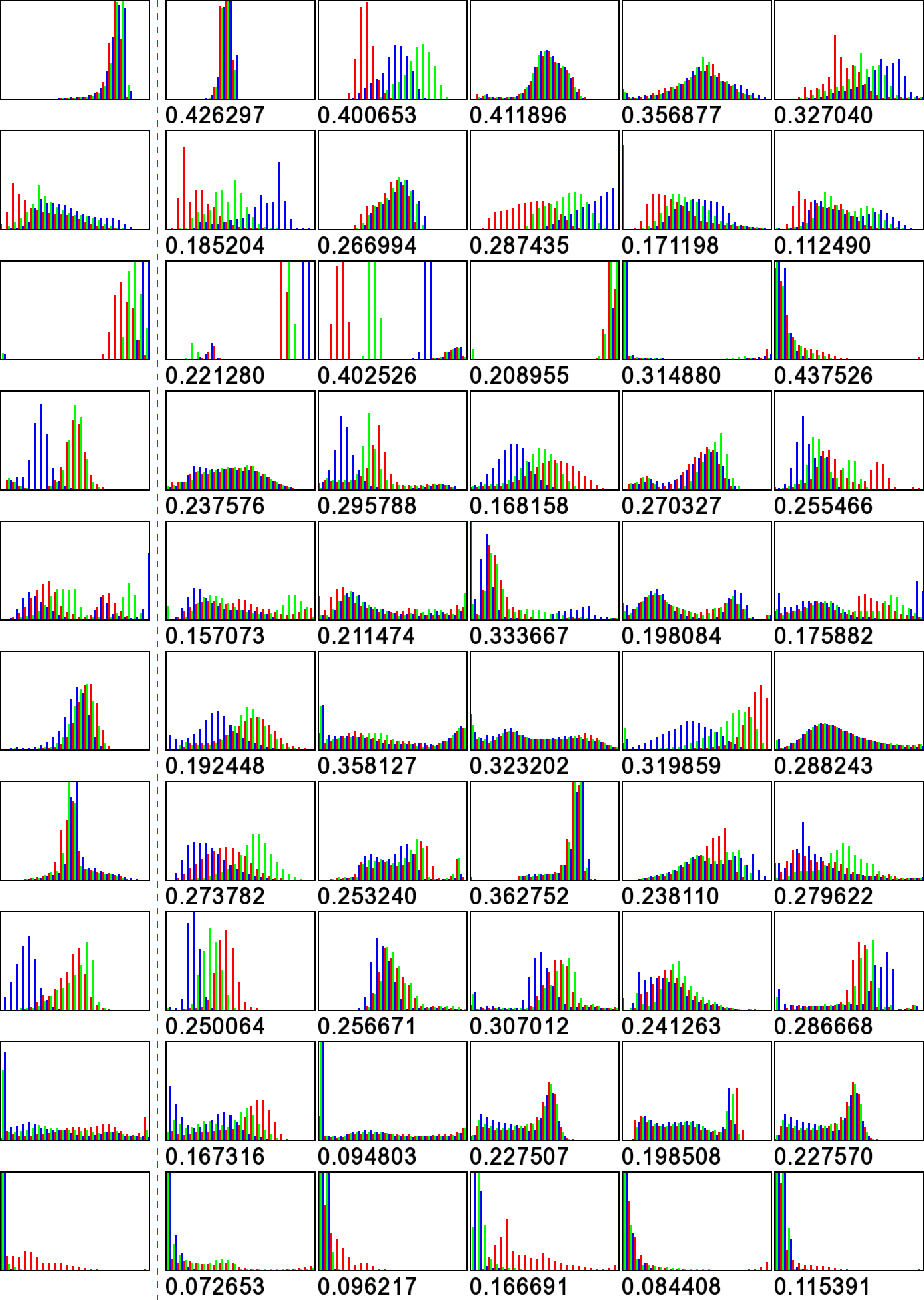}
\caption{Histograms of colors for the top-5 images presented in Fig. \ref{fig:sim_base}.}
\label{fig:hist_base}
\end{figure}

\begin{figure}[!t]
\centering
\includegraphics[width=3.0in]{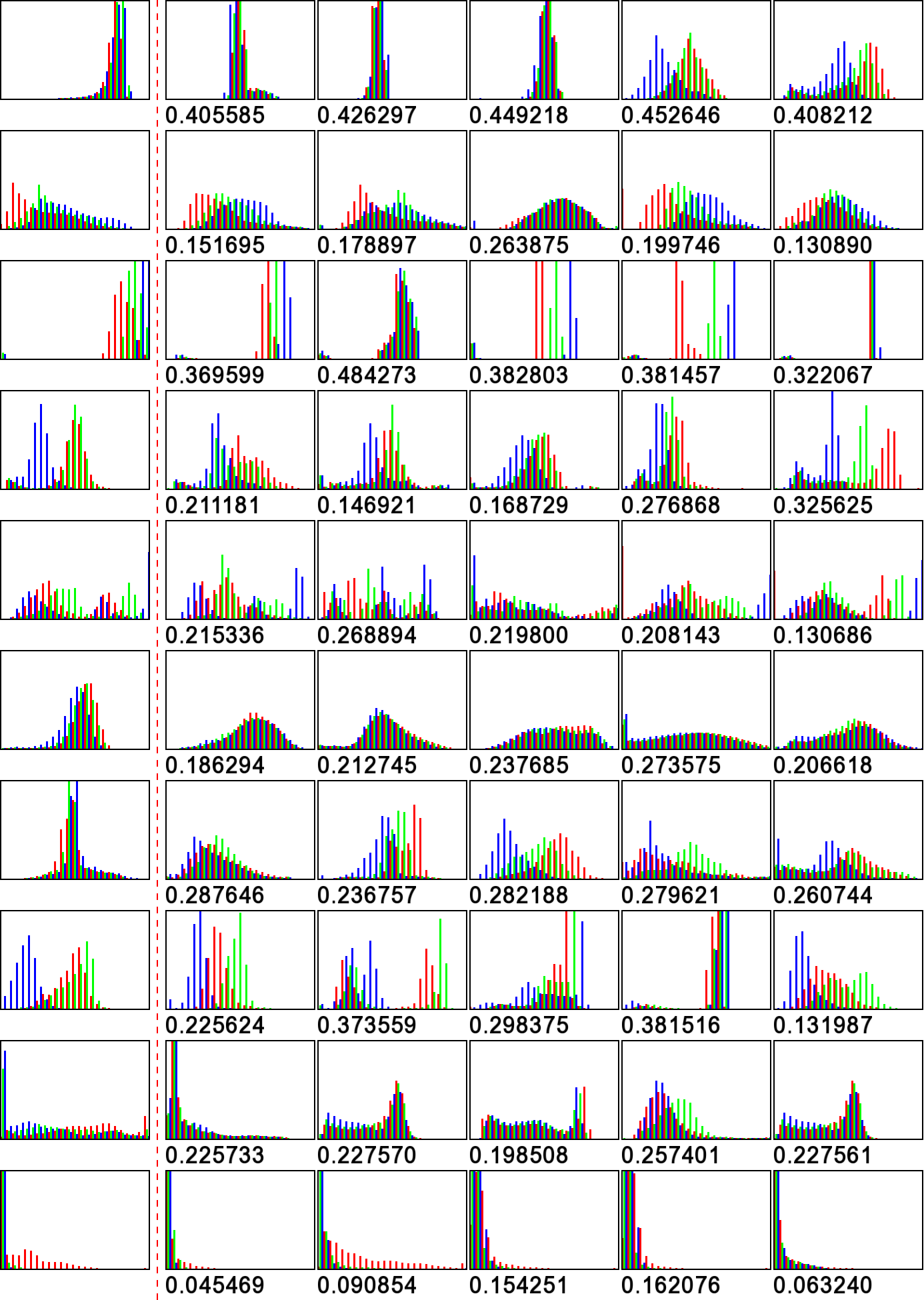}
\caption{Histograms of colors for top-5 images presented in Fig. \ref{fig:sim_fc}.}
\label{fig:hist_fc}
\end{figure}

\begin{figure}[!t]
\centering
\includegraphics[width=3.0in]{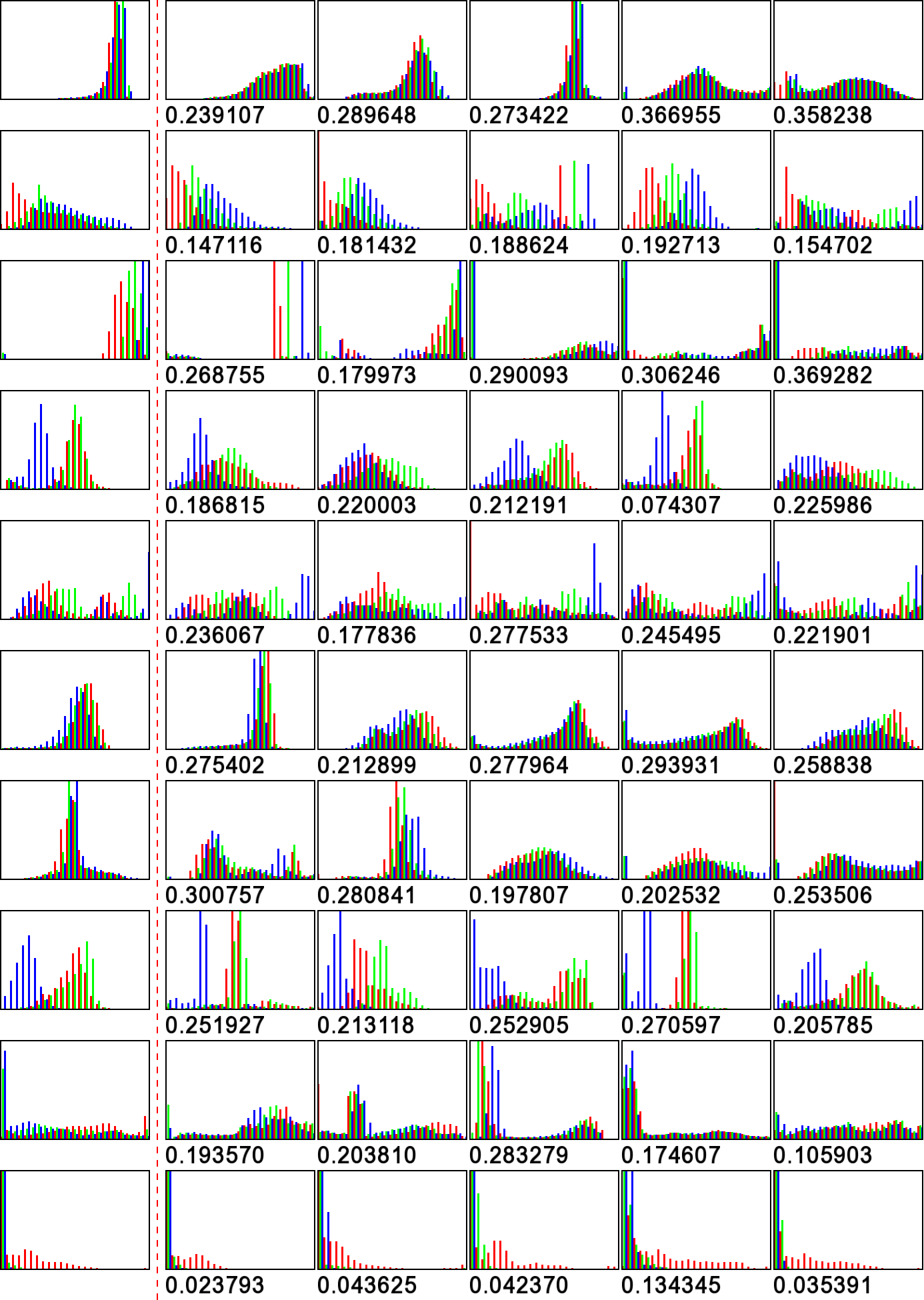}
\caption{Histograms of colors for top-5 images presented in Fig. \ref{fig:sim_cnn}.}
\label{fig:hist_cnn}
\end{figure}

\begin{figure}[!t]
\centering
\includegraphics[width=3.0in]{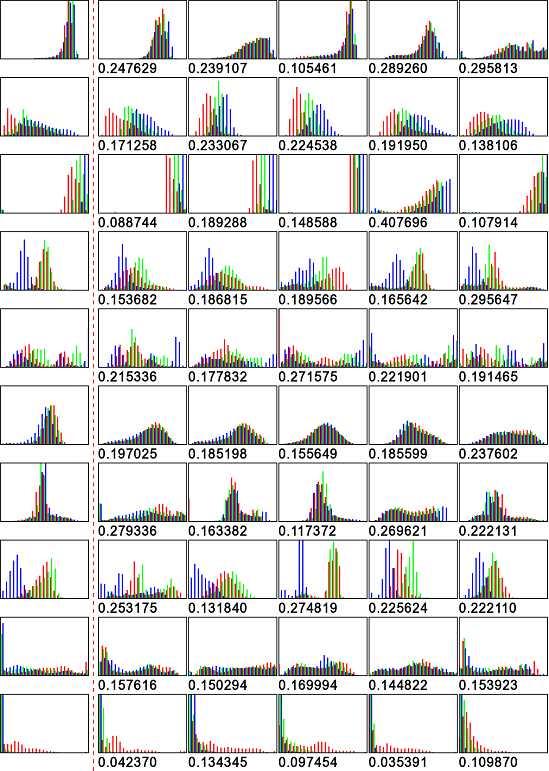}
\caption{Histograms of colors for top-5 images presented in Fig. \ref{fig:sim_own}.}
\label{fig:hist_own}
\end{figure}

As we already pointed out, the quantitative assessment of the obtained results is problematic. However, to prove that the returned images for the $\eta(x)$ descriptors match better to the query image with respect to textures, background, etc., we decided to analyze the color histograms. For each considered image we generated three histograms (one for each of the RGB colors). For each color we divided the range of possible values, i.e. $[0;255]$, into $25$ bins, each covering approximately the interval of $10$ values. The results obtained for images shown in Figures \ref{fig:sim_base}-\ref{fig:sim_own} are presented in Figures \ref{fig:hist_base}-\ref{fig:hist_own}, respectively. 

As can be observed visually, the histograms concerning images retrieved using the $\eta(x)$ descriptors match the best to the histograms for the query images. Each histogram can be treated as a vector (with the number of dimensions three times larger than the number of bins). Hence, the histograms can be compared as vectors, using, for example, the $L_{1}$ distance. Below each histogram in Figures \ref{fig:hist_base}-\ref{fig:hist_own} we placed a number which is the $L_{1}$ distance between the considered image and the corresponding query image, divided by the number of dimensions. The presented quantitative results also confirm that the application of the $\eta(x)$ descriptors gives the best results of image retrieval, assuming that we care about the similarity of characteristics like background or color distribution as well as about the class compatibility. The obtained results are additionally summarized in Tab. \ref{tab1}, which contains the values of histogram distances for each query image averaged over all $5$ retrieved images. As can be seen, descriptors $\eta(x)$ based on the information from both convolutional and fully connected layers provide the best results in $6/10$ cases. It is worth noticing that descriptors $\tilde{\eta}(x)$, based solely on the activations from the convolutional layers, win for three query images and in many other cases they take the second place. This fact confirms once again that the use of information taken from the convolutional part of the deep network, extracted in a proper manner, can help in image retrieval task, assuming that the user cares about characteristics like background or textures.

\begin{table}[htbp]
\caption{Average values from histogram results per classes}
\begin{center}
\begin{tabular}{|c|c|c|c|c|}
\hline
\textbf{Query} &
\textbf{IMAGENET1M} &
\textbf{FC} &
\textbf{Conv.} &
\textbf{FC + Conv.}\\
\textbf{Image} & $IM(x)$ & $h(x)$ & $\tilde{\eta}(x)$ & $\eta(x)$\\
\hline
\textbf{1} & 0.384553 & 0.428392 & 0.305474 & \textbf{0.235454}\\
\textbf{2} & 0.204664 & 0.185021 & \textbf{0.172917} & 0.191784\\
\textbf{3} & 0.317033 & 0.388040 & 0.282870 & \textbf{0.188446}\\
\textbf{4} & 0.245463 & 0.225865 & \textbf{0.183860} & 0.198270\\
\textbf{5} & 0.215236 & \textbf{0.208572} & 0.231766 & 0.215622\\
\textbf{6} & 0.296376 & 0.223383 & 0.263808 & \textbf{0.192215}\\
\textbf{7} & 0.281501 & 0.269391 & 0.247089 & \textbf{0.210368}\\
\textbf{8} & 0.268336 & 0.282212 & 0.238866 & \textbf{0.221514}\\
\textbf{9} & 0.183141 & 0.227355 & 0.192234 & \textbf{0.155330}\\
\textbf{10} & 0.107072 & 0.103178 & \textbf{0.055905} & 0.083886\\
\hline
\textbf{Average} & 0.250337 & 0.254141 & 0.217479 & \textbf{0.189289}\\
\hline
\end{tabular}
\label{tab1}
\end{center}
\end{table}

\newpage
\section{Conclusions and future research directions}
\label{sec5}
In this paper, we proposed new descriptors for image retrieval, based on the neural activations of deep neural networks. The descriptors contain information from activations of both convolutional and fully connected layers of the network. Thanks to this solution the images retrieved from the dataset not only match to a query image semantically but also suit concerning other characteristics, like color distribution, textures or image background. The effectiveness of the proposed descriptors was demonstrated in numerical experiments conducted on the IMAGENET1M dataset. The applied deep structure was the VGG16 neural network. Despite achieving very promising results -- in most cases superior to the previously developed algorithms in the literature -- the descriptor construction algorithm presented in this paper opens a wide area of possible research and can be further developed in many ways. One of the possible paths is to apply more sophisticated ways of computing particular descriptor elements. In this paper, we proposed to use arithmetic or weighted averages over significant activations of the whole feature map, which may be too general. Another possibility for future research is to replace the VGG16 by the model with higher classification efficiency. Then we can suspect the image comparison quality to be also significantly better. Moreover, the classification model could be replaced by the neural network designed for semantic segmentation. In the VGG16 net, despite using the activations from convolution layers in proposed descriptors, all information contained in the model is mainly focused on the characteristics of the class, which is assigned to the considered image in a one-to-one manner. On the contrary, the semantic segmentation model analyzes each pixel of the image and assigns it to a specific class of the object contained in the image. The application of such a model would allow the semantic information  analysis, oriented simultaneously on image class as well as on the environment surrounding the object on this image. The construction of a descriptor based on a semantic analysis model will be investigated in our future work.

\ifCLASSOPTIONcaptionsoff
  \newpage
\fi

\bibliographystyle{IEEEtran}
\bibliography{IEEEabrv,references}

\end{document}